# A Compendium of Autonomous Navigation using Object Detection and Tracking in Unmanned Aerial Vehicles

Mohit Arora[1], Pratyush Shukla[2], Shivali Chopra[3]

School of Computer Science and Engineering

Lovely Professional University. Phagwara, Punjab, India

[1]mohit.15980@lpu.co.in, [2]pratyush.11708251@lpu.in, [3]shivali.19259@lpu.co.in

**Abstract**

Unmanned Aerial Vehicles (UAVs) are one of the most revolutionary inventions of 21$^{st}$ century. At the core of a UAV lies the central processing system that uses wireless signals to control their movement. The most popular UAVs are quadcopters that use a set of four motors, arranged as two on either side with opposite spin. An autonomous UAV is called a drone. Drones have been in service in the US army since the 90's for covert missions critical to national security. It would not be wrong to claim that drones make up an integral part of the national security and provide the most valuable service during surveillance operations. While UAVs are controlled using wireless signals, there reside some challenges that disrupt the operation of such vehicles such as signal quality and range, real time processing, human expertise, robust hardware and data security. These challenges can be solved by programming UAVs to be autonomous, using object detection and tracking, through Computer Vision algorithms. Computer Vision is an interdisciplinary field that seeks the use of deep learning to gain a high-level understanding of digital images and videos for the purpose of automating the task of human visual system. Using computer vision, algorithms for detecting and tracking various objects can be developed suitable to the hardware so as to allow real time processing for immediate judgement. This paper attempts to review the various approaches several authors have proposed for the purpose of autonomous navigation of UAVs by through various algorithms of object detection and tracking in real time, for the purpose of applications in various fields such as disaster management, dense area exploration, traffic vehicle surveillance etc.

# 1. Introduction

Unmanned Aerial Vehicle (UAV) is an airplane sans human pilot. They are a piece of Unmanned Aerial System (UAS) that additionally incorporates a ground-based controller and a system of correspondences between the UAV and the controller. UAVs originated for use in the military but have been finding wide acceptance in commercial, agricultural, scientific, recreational and miscellaneous applications such as aerial photography, journalism, drone racing etc. As of 2019, civilian UAVs have surpassed military UAVs, with an estimate that the current trend will increase by over a billion till 2025. Issues such as ground controller setup, secure protocol for signal transmission, real time video feed, prolonged flight time, skilled piloting[1] are some of the major challenges that need to be addressed while deploying a UAV outside. An autonomous UAV, called a drone, addresses the challenge of skilled pilot by allowing the drone to navigate itself in a given area. By using computer vision and deep learning, we make progress in developing algorithms that can detect and track objects in real time[2] [3] without first scanning the area such that image processing the drone's system hardware is capable of image processing. UAVs are equipped with a fairly complex hardware that has decent processing speed but is limited in use due to use of CISC based ARM processors. The system hardware is called a flight controller and is usually a System on a Chip (SOC) that controls the components of the UAV. Due to the power required for lifting a UAV, typically Lithium Polymer batteries are used which have high power output but lack the flexibility provided by Lithium Ion batteries. As a result, designing a drone with high computing power requires large battery capacity which limits their flight time and is thus, ineffective for image processing on board. Hence, designing algorithms that are efficient and robust without requiring high computing power [as those provided by Graphical Processing Units (GPUs)][4] [5] becomes a requirement. Various algorithms based on Computer Vision and Deep Learning using pretrained models have been proposed and developed and are available as open source libraries[6]. Several algorithms require specific camera hardware so that they perform well their tasks. We attempt to present some of the research done in the field of autonomous flight of UAVs using object detection and tracking[7], in complex environments, that are accurate and efficient without compromising their system hardware[8]. The research paper is presented as following – Section 2 highlights the motivation behind the research work, Section 3 presents the relevant literature survey undertaken in the course of research and Section 4 provides the conclusion and future scope of the work.

## 2. Motivation

Our main motivation is to present the opportunities that the field of UAV and Computer Vision present today with the availability of such hardware that can compute billions of numbers in seconds. Algorithms can be developed to detect different kinds of objects, the information gained can be used for in-depth insights and the purpose that it may fill. This reduces dependency on external modules like sensors and boards that should be attached and programmed as such which can only detect object but not infer their kind, whether it may pose as an obstacle or can be a threat to the vehicles. We thus present this paper to shed light on the importance of using computer vision in UAVs so that autonomous navigation in different environments for commercial, scientific, research purpose is economical, robust and accurate.

## 3. Related Literature

Lynne Grewe et.al [2019][9] designed a system Seeing Eye Drone for the visually impaired by providing environmental awareness and object detection. This system compares 3D and 2D modalities of vision on a UAV by using a system of two networks – the first one is an SSD CNN architecture that calculates target's direction and travel location while the second network identifies location of familiar objects using an Android application. The results are communicated by Text-To-Speech to the visually impaired. The accuracy of this system is 50% when detecting an object and 86% when classifying it as an object of interest in case the object is detected. It can be improved by using improved pre-trained generalized object detector model of CNN, higher amount of cleaned and assimilated data and better-quality sensors.

M. Alwateer et. al [2019][10] provided a review of drone applications, its data management and services, distributed computing, and human-drone interaction. It highlights the issues associated with them such as data management and system design, Big Data, optimization configuration of ground and airborne infrastructure for best QoS & QoE etc. They found various applications for the drones such as delivery, filming, tourism, helping of disabled, surveillance, disaster management, mapping and agriculture while also claiming autonomous drones are much popular for such uses. The issues encountered are as follows – data mules, where data collection using drones in widely deployed networks is a challenge as it must satisfy constraints relating temporal and spatial communication range. As drones have cameras and various sensors built in them, they collect large heterogenous data that needs to be analyzed and stored securely for tamper proofing.

The paper provides an insight in the various applications and challenges a drone can have such that their end use can be enhanced and purpose be suitable in various areas.

Maciel-Pearson et. al [2019][11] proposed a learning approach based on end-to-end multi-task regression that uses deep learning for autonomous drone control in outdoor environments irrespective of the presence of various sensors with shallow networks as their advantage to individually learn each task. They claimed their approach beats the other state-of-the-art techniques, can cover wider regions, is able to generalise previously unexplored environments and can densely explore within a specific perimeter. Data was collected using the FrSky Taranis Plus Digital Telemetry Radio System. The training was performed on AirSim software using GTX 1080Ti and Intel Xeon processors. They conclude that their proposed approach is best suitable for search and rescue operations for it does not require GPS navigation or unique pathways for navigation.

Trung-Thanh et. al [2019][12] performed the impact analysis of zooming capabilities of various lens types attached to a drone for operations related to search and rescue. UAV's are embedded with sensors, camera, additional computing boards etc. along with miscellaneous items such as emergency health items, medical supplies during SAR operations which limits their flight times. Often the camera systems have low image quality which leads to the UAV having to move closer to the target hence consuming more battery and energy. The above proposed system tends to increase the flight time by contributing to the latter factors mentioned. An algorithm is proposed by the authors to calibrate the camera zoom lens intelligently. The paper concludes that this system prolongs the battery life of the drone by decreasing the travel distance and the flight time.

Minjie Wan et. al [2019][13] proposed a target tracking system for UAV in intelligent urban surveillance systems for applications in smart city. They propose an algorithm using sparse representation or sparse coding theory wherein a target sample is sparsely placed in a subspace over a joint template dictionary constrained by L1 regularization then coefficient of sparse representation is constrained further using L2 regularization on the basis of temporal consistency and finally, a binary support vector having contiguous occlusion constraint which is exploited to balance the occlusion occurring partially in UAV videos. For long range tracking, the algorithm is implemented using particle filter and dynamically updating the temporal dictionary on the basis of average cosine similarity to cope up with the change in appearance of target. Their experimentation shows they were able to achieve satisfactory performances in various UAV scenes considering accuracy and robustness.

Jianxiu Yang et. al [2019][14] suggest that vehicle detection using UAVs is a captivating task due to small size of objects, complex background and imbalanced vehicle image samples They thus propose a high-performance vehicle detector for UAV. The proposed model contains a single shot refined neural network that provides contextual information using top-down architecture. They claim an accuracy of 92% and 90.4% on their collected dataset and the publicly available Stanford drone dataset while also achieving real time detection.

Minjung Lee et. al [2019][15] suggest that the calibration pattern of a fisheye camera works on a 2D input scene when it is in a prespecified position while the lens design specification is under the expertise of optical experts. The fisheye lens has geometrical distortions that include barrel and tangential distortions. They propose an algorithm based on image correction that rectifies the geometric distortion and claim that it can be used with a Virtual Reality[16] camera with 195° field-of-view (FOV) pair of fisheye lenses. The proposed algorithm works in 3 stages – detection of features, estimation of parameter of distortion and selection of optimal corrected image from the various corrected images. Though such lens provides a wide field-of-view, they have limited practical application due to non-linear distortion of the captured image with an added expense of complicated pre-processing procedure to correct the distortion.

Suet-Peng Yong et. al [2018][17] suggest an algorithm for human object detection using deep learning and computer vision for human detection in forest areas for the application of reducing illegal entries and illegal forest activities while also saving time and cost. Hence, they apply their algorithm in the area of forest surveillance. They used 3DR Solo Drone with a GoPro Hero 4 attached for image data collection and feeding them in MobileNet architecture with Single Shot MultiBox Detector for object detection which is faster than traditional machine learning techniques as it leverages the power of both CPU and GPU in real time. An accuracy of 97.3% was achieved.

Okan Erat et. al [2018][18] investigate the potential of UAV-augmented human vision system which is a mixed reality system for exocentric control of the UAV. The proposed system projects the live streamed feed of the camera drone onto the user's display in a 3D exocentric space for user's comfort. They also implemented high level experimental interaction techniques to indirectly control the drone. The proposed system, while technically feasible, needs more progress to achieve the desired goal while also addressing the technical challenges that may come with it.

Kathiravan Natarajan et. al [2018][19] proposed and developed a computer vision-based library that uses a drone's camera for capturing images of surrounding then use pattern recognition for acquiring meaningful information from the image data. Their proposed framework involves segregation of image from real time streams of video from front camera of the drone, resulting in robust and trustworthy image recognition based on the images segregated and conversion of gestures classified from the images into gestures based on drone movements such as takeoff, landing, hovering etc. The proposed algorithm uses Haar features in AdaBoost classifier for gesture recognition. The accuracy of the proposed algorithm is 90% at a distance of 3 feet.

F. Valenti1 et. al [2018][20] proposed a system based on computer vision for autonomous landing of UAVs in environments where GPS fail. An omni-directional stereo-vision setup was used to render image data into a three-dimensional version of the environment. A local obstacle grid is managed using various frames and adjusted in accordance with the Unmanned Aerial Vehicle's locomotion. The problem of self-localization was solved using Simultaneous Localization and Mapping algorithm. There proposed system has challenges such as memory inefficiency when updating the environment which can be solved using some caching strategy and noisy mapping of environments due to the limitation of stereo matching algorithm but can be challenged using different sensing algorithms.

Jasmin James et. al [2018][21] propose a new technique for long range aircraft detection using CNN for visual features identification and learning, of aircraft using the aircraft data. The state-of-the-art hand-crafted features are fused with these learned features and compared against on spot data captured from Unmanned Aerial Vehicles. The CNN uses modified SegNet architecture for aircraft detection with morphological processing. The proposed system achieves mean detection range improvement of 299m without any false alarms.

Wei Zhang et. al [2018][22] suggests that a target's aspect ratio changes over time during a UAV's tracking phase which makes it challenging. They introduced a fine coarse tracking system for handling this issue as traditional trackers maintain a certain aspect ratio. They use a coarse tracker that dynamically works with the dynamic aspect ratio along with a fine tracker that adjusts the boundaries of the tracking object. Their system outperforms existing trackers with a significant accuracy.

Jeonghoon Kwak1 et. al [2018][23] suggested a graph-based path for short routes for UAV autonomous flight[24] [25] using A* search algorithm to capture images at multiple surveillance positions. It is done in 3 phases – Record, Graph, and Path Planning Phase, the pseudocode for which is available in their research paper. The proposed method was experimentally verified using 8115 collected flight points with 109 being selected. Along the 6 flight times, the recorded flight path was 1364.32m while the flight path obtained using the proposed method was 764.27m which lowered the granularities of created graphical paths to 1.34%.

Sushil Pratap Bharati et. al [2018][26] propose a real time obstacle detection and tracking technique integrated with Kernalized Correlation Filter (KCF) framework. They use this technique to refine the boundary and object's location whenever the accuracy of tracker drops down a predefined threshold. They also implement a technique of post-processing to correctly localize the obstacle recovered from the searched region from a saliency map. They claim their experiment outruns the state-of-the-art techniques significantly on the basis of speed and accuracy.

Jiasong Zhu et. al [2018][27] present an advanced urban traffic density estimation elucidation by capturing very high-quality video of traffic vehicles using UAVs. Their approach of hybrid Deep Learning Network detectors outperform other Deep Neural Network-based approaches. Their solution is robust, accurate and efficient.

The consolidated literature review has been tabulated in Table 1 given beneath.

Table 1. Literature Review Summary

| Paper title | Architecture | Dataset | Research Findings |
|---|---|---|---|
| Grewe, Lynne et. al [9] | COCO Mobile Net SSD CNN | COCO, User Heading | Accuracy is 50% during object detection and 86% when classifying objects. |
| Alwateer, M. et. al [10] | Various | Various | Drones can be multipurposed. With each new application area, new challenges have to be solved accordingly. |
| Maciel Pearson et. al [11] | LSTM (Shallow Network) | Sourced manually | A novel approach using Regression-based learning bypassing use of GPS entirely. Work needs to be done to enhance approach in limited navigational space. |
| Trung Thanh et. al [12] | DJI A2 Flight Controller | Sourced manually | Demonstrated use of an automatic superzoom digital lens camera system with automatic lens calibration algorithm. It prolongs drone flight time. |
| Wan, Minjie et. al [13] | Artificial Neural Network with MRF-based binary support vector | TB50 & TB100 | Uses algorithm for target tracking based on sparse representation without using pretrained models. Can detect and track objects in real time even though UAV videos suffer from noise and disturbances. |
| Yang, Jianxiu et. al [14] | RefineDet | Stanford and manually collected | Accuracy of 92% on collected dataset and 90.4% on Stanford dataset. This technique detects small-sized vehicles in real time more accurately. |
| Lee, Minjung et. al [15] | Samsung Gear 360 Camera | Sourced manually | Demonstrated algorithms for barrel distortion correction in fisheye lens camera that can be used as a pair in VR or AR. Camera system can have wide field-of-view thus extending application in drones. |
| Yong, Suet Peng et. al [17] | MobileNet and SSD | Sourced manually | Using deep learning framework, human detection in forest areas is easy through this technique though practical approach based on drone hardware needs more work. |

| Erat, Okan et. al [18] | PX4 Pixfalcon autopilot with Odroid XU3 single-board processor (SBC) | Sourced manually | Developed an experimental hardware system using Microsoft HoloLens and an advanced drone hardware. An expensive setup that indicates how exocentric view using drones can be implemented |
|---|---|---|---|
| Natarajan, Kathiravan et. al [19] | Artificial Neural Networks using Haar feature based AdaBoost classifier | Sourced manually | Specifically developed system called Parrot AR.Drone 2.0 is used which limits the use to high end drone systems. Hardware capability required is high. |
| Valenti, F. et. al [20] | Simultaneous Localization and Mapping (SLAM) with Extended Kalman Filter (EKF) | Sourced manually | Environments where GPS accuracy is low allows this approach to work better but with the use of specific hardware. A more accurate technique can be developed using existing approaches. |
| James, Jasmin et. al [21] | SegNet | Project ResQu dataset and Project Smart Skies dataset | Semantic Pixel-wise segmentation done using CNN. |
| Zhang, Wei et. al [22] | End-to-End Reinforced Learning Architecture | UAV123 and VisDrone | Coarse-to-fine tracker proposed tackles aspect ratio problem of UAV drone videos and gains extensive improvement in other challenges such as occlusion, SV etc. |
| Kwak, Jeonghoon et. al [23] | DJI Inspire | Sourced manually | Uses graph-based approach for autonomous navigation of UAV using A* search algorithm instead of computer vision |
| Bharati, Sushil Pratap et. al [26] | Salient Object Detection with Kernalized Correlation Filter (KCF) | 25 challenging video sequences sourced manually | Demonstrated technique tackles all the challenges of UAV videos such as occlusion, camera instability, noise etc. |

| Zhu, Jiasong et. al [27] | Deep Vehicle Counting Framework (DVCF) with Enhanced SSD | UavCT | Technique integrates deep learning and computer vision approaches and is robust and accurate in detecting vehicles. Allows flexibility to detect other type of objects. |

4. Conclusion and Future scope

Progress made in the field of computer vision has been immense. Advanced algorithms that can detect and track various type of objects or classify them accordingly with improved accuracy and agility are still being developed. Since UAVs are equipped with camera equipment to provide visual feed to the user controlling the drone for environmental awareness, algorithms can be developed suitable to the system hardware for aforementioned purpose. Though some challenges are present due to power limitation that limit the use of advanced powerful hardware, they can be tackled by developing powerful algorithms.